\documentclass{article}
\usepackage{spconf,amsmath,graphicx,amsfonts}
\usepackage{cite}
\usepackage{booktabs} 
\usepackage{tabularx}
\usepackage{enumitem}
\usepackage{textcomp}
\usepackage{cite}
\usepackage[english]{babel}
\usepackage{mathtools}

\usepackage{ifpdf}
\usepackage{algorithmic}
\usepackage{multirow}
\usepackage{xcolor}
\usepackage{array}
\usepackage{url}
\usepackage[colorlinks=true, urlcolor=blue, linkcolor=red]{hyperref}
\usepackage{comment}

\title{Taxes are All You Need: Integration of Taxonomical Hierarchy Relationships into the Contrastive Loss}
\name{Kiran Kokilepersaud \enskip Yavuz Yarici \enskip Mohit Prabhushankar \enskip Ghassan AlRegib \enskip }

\address { OLIVES at the Center for Signal and Information Processing CSIP,\\ 
School of Electrical and Computer Engineering, Georgia Institute of Technology, Atlanta, GA, USA \\
\{kpk6, yavuzyarici, mohit.p, alregib\}@gatech.edu             }
\begin{document}
%

\twocolumn[{%

{ \large
\begin{itemize}[leftmargin=2.5cm, align=parleft, labelsep=2cm, itemsep=4ex,]

\item[\textbf{Citation}]{K. Kokilepersaud,Y. Yarici, M. Prabhushankar, G. AlRegib, "Taxes Are All You Need: Integration of Taxonomical Hierarchy
Relationships into the Contrastive Loss," in \textit{2024 IEEE International Conference on Image Processing (ICIP), Abu Dhabi, United Arab Emirates (UAE), 2024.}}

\item[\textbf{Review}]{Date of Acceptance: June 6th 2024}

\item[\textbf{Codes}]{\url{https://github.com/olivesgatech/TaxCL}}

\item[\textbf{Bib}]  {@inproceedings\{kokilepersaud2024taxes,\\
    title=\{Taxes Are All You Need: Integration of Taxonomical Hierarchy
Relationships into the Contrastive Loss\},\\
    author=\{Kokilepersaud, Kiran and Yarici, Yavuz and Prabhushankar, Mohit and AlRegib, Ghassan and Corona, Enrique and Singh, Kunjan and Parchami, Armin\},\\
    booktitle=\{IEEE International Conference on Image Processing\},\\
    year=\{2024\}\}}


\item[\textbf{Contact}]{
\{kpk6, yavuzyarici, mohit.p, alregib\}@gatech.edu\\\url{https://ghassanalregib.info/}\\}
\end{itemize}

}}]

\maketitle
\begin{abstract}
In this work, we propose a novel supervised contrastive loss that enables the integration of taxonomic hierarchy information during the representation learning process. A supervised contrastive loss operates by enforcing that images with the same class label (positive samples) project closer to each other than images with differing class labels (negative samples). The advantage of this approach is that it directly penalizes the structure of the representation space itself. This enables greater flexibility with respect to encoding semantic concepts. However, the standard supervised contrastive loss only enforces semantic structure based on the downstream task (i.e. the class label). In reality, the class label is only one level of a \emph{hierarchy of different semantic relationships known as a taxonomy}. For example, the class label is oftentimes the species of an animal, but between different classes there are higher order relationships such as all animals with wings being ``birds". We show that by explicitly accounting for these relationships with a weighting penalty in the contrastive loss we can out-perform the supervised contrastive loss. Additionally, we demonstrate the adaptability of the notion of a taxonomy by integrating our loss into medical and noise-based settings that show performance improvements by as much as 7\%.
\end{abstract}

\section{Introduction}

\begin{figure}[ht]
\centering
\includegraphics[scale = .2]{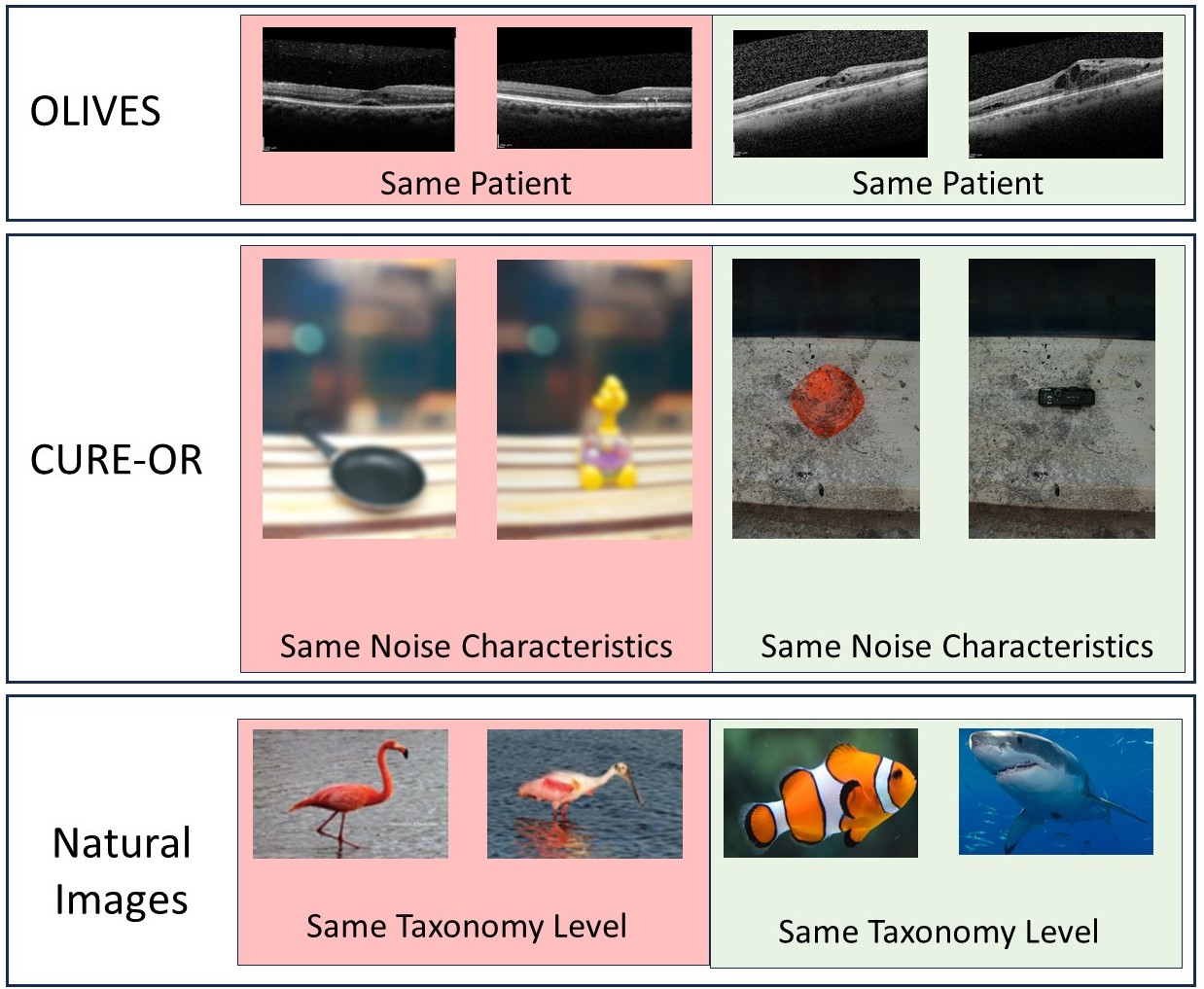}
\caption{Every dataset has semantic dependencies that exist beyond just the basic task label. This shows examples on the datasets OLIVES \cite{prabhushankar2022olives}, Cure-OR\cite{temel2017cure}, and on natural images. \vspace{-.6cm}}
\label{fig: tax}
\end{figure}

\begin{figure}[ht]
\centering
\includegraphics[scale = .17]{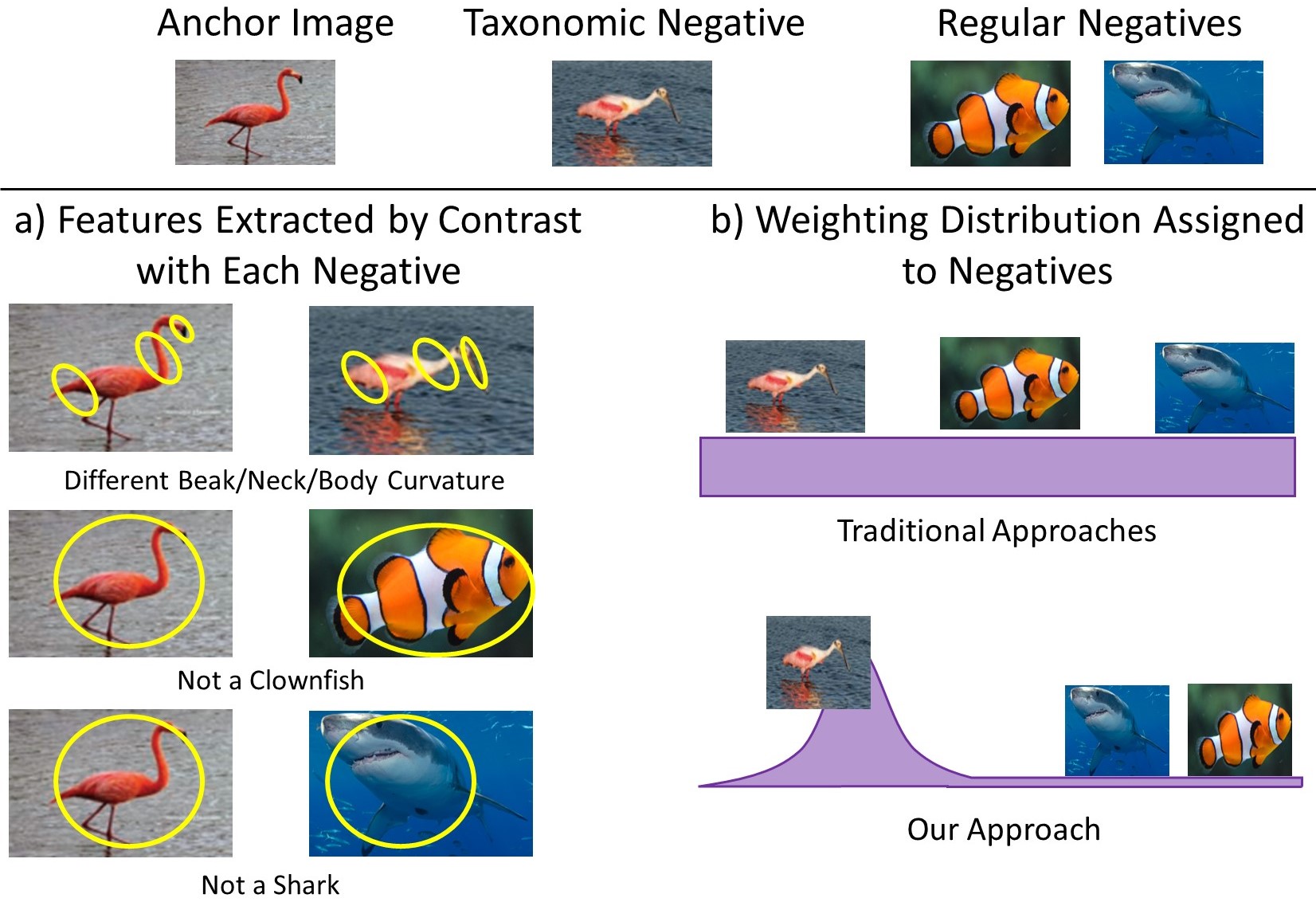}
\caption{a) When contrasting against negatives within the same taxonomy, the model is forced to learn more fine-grained differentiating features in order to rectify the difference between the two. b) Traditional approaches treat all negatives equally and do not consider the importance of the taxonomic negatives. We address this through additional importance weighting on taxonomic negatives. }
\label{fig: batch_view}
\end{figure}

Contrastive learning \cite{le2020contrastive} refers to a family of self-supervision strategies that builds an embedding space where similar pairs of images (positives) project closer together and dissimilar pairs of images (negatives) project apart. In the supervised setting \cite{khosla2020supervised}, these positive and negative sets are selected on the basis of an associated class label for the downstream task of interest. The advantage of this approach over supervised learning algorithms that operate on an output probability space, such as the cross-entropy loss, is that it directly penalizes the structure of the representation space itself. While this setup has exhibited great empirical success, the simple class-wise assignment of positive and negative sets does not consider other semantic relationships that commonly exist within data. Consider the example shown in Figure \ref{fig: tax}. Within each dataset, there are higher order semantic structures that describe relationships between classes. This includes patient relationships, similar noise characteristics, and memberships of the same species grouping (i.e. birds and fish). This multi-level hierarchical organization of semantic relationships is known as a taxonomy. Work has suggested \cite{bengio2013representation} that representations that reflect a hierarchical organization of factors of variation within the data are more likely to generalize well to downstream tasks. Further intuition can be built through the example shown in part a) of Figure \ref{fig: batch_view}. We see that taxonomic negatives oftentimes contain informative features, such as bird-like features, in common with the anchor image. Thus contrast against a taxonomic negative forces the model to consider more fine-grained differentiating features that can help to better distinguish the semantic category of the anchor image~\cite{9190927}. This motivates the incorporation of additional taxonomic structures beyond just the class label into a contrastive representation learning framework. However, the standard supervised contrastive loss does not have any way to introduce this additional structure into the representation space.

In this work, we introduce a novel supervised contrastive loss that directly integrates semantic taxonomy labels into the loss function. This is accomplished by decomposing the negative distribution into two subsets: taxonomy negatives and regular negatives. We then introduce an additional reweighing function on top of taxonomy negatives that effectively biases the loss to introduce additional spread with respect to these samples over regular negatives. This results in a negative distribution that is skewed towards samples of the same taxonomic grouping as shown in part b) of Figure \ref{fig: batch_view}. In this way, the negative distribution now reflects the structure of semantic relationships that exist with respect to each individual instance in a batch. We also perform an analysis that shows this additional bias is helpful due to properties that samples of taxonomic groupings exhibit: dimensional collapse \cite{jing2021understanding} and greater cosine similarity with the anchor image embedding.
In summary, we make the following contributions:
\begin{enumerate}
    \item We propose a novel supervised contrastive loss called TaxCL that can effectively integrate taxonomic semantic relationships.
    \item We show that the idea of a taxonomy can be extended within a variety of data settings such as noise and clinical information. 
    \item We demonstrate that our approach out-performs variants of the supervised contrastive (SupCon) loss and shows further improvement when used as part of linear combination of losses with SupCon. 
    \item We analyze this interaction of different taxonomic levels through an eigendecomposition and cosine similarity study of the representation space.
\end{enumerate}

\label{sec:intro}

\section{Related Work}

\subsection{Positive and Negative Set Relationships in Contrastive Learning} 

Contrastive learning algorithms are defined by how positive and negative sets are defined as well as how the algorithm influences the relationship between the two. Research has primarily revolved around the idea of manipulating a defined set of positives and negatives based on augmentations, label information, or negative queues \cite{chen2020simple,khosla2020supervised}. A subset of this work has involved adapting positive selection towards specific application domains such as medicine\cite{kokilepersaud2022clinical}, fisheye object detection, \cite{kokilepersaud2023exploiting}, and associating seismic structures \cite{kokilepersaud2022volumetric}. More general work has included ideas such as removing the influence of false negatives \cite{huynh2022boosting} and creating new types of representations \cite{lavoie2022simplicial}. There is also a set of works that revolve around the concept of exploiting the relationship between the positive set and the negative set. This includes introducing hardness estimation \cite{robinson2020contrastive}, selecting negative samples based on a ring of distances from the anchor point \cite{wu2020conditional}, and making use of inter-anchor relationships \cite{zhang2022dual}. While these works are similar to ours in the sense of exploiting positive and negative relationships, they do not consider the problem of directly integrating different levels of semantic taxonomic information.

\subsection{Hierarchies and Taxonomies in Contrastive Learning}

A variety of contrastive learning approaches have been proposed to introduce some notion of hierarchical relationships into the process of shaping representations. However, oftentimes, this ``hierarchy" is with regards to the physical structure of the task of interest, rather than with respect to semantic relationships that may exist in the data. \cite{huang20223} showed how to make use of different levels of feature importance within a 3D transformer for hyperspectral image classification. \cite{feng2023hierarchical} formulated different levels of transformer features as part of hierarchical restoration of corrupted images. Other work has attempted to estimate semantic relationships through various types of clustering strategies. \cite{lin2022contrastive} demonstrated that a hyperbolic space with a contrastive loss can mimic a hierarchical structure. \cite{gui2022improving} demonstrated the usage of clustering with a smaller number of supervised samples to estimate a superclass structure.  There has also been work that attempts to use a taxonomic structure, but does not directly integrate it within a contrastive loss. \cite{hu2022hiure} enforces a penalty across a linear weighted sum of contrastive losses associated with each semantic label in a multi-label setting. While all of these works make use of some hierarchical structure, they are in contrast to our work where we reformulate the contrastive loss itself to directly influence the positive and negative relationship based on the taxonomy information present. We show that this reduces the dependency on application specific structures and allows the integration of a wide variety of different types of ``taxonomies."
\section{Representation Analysis}

Our goal is to integrate semantic taxonomy information into the supervised contrastive loss. This semantic information provides additional context towards shaping the structure of the representation space. While these advantages seem intuitive, it isn't clear how these benefits manifest on top of a representation space. Therefore, to motivate our proposed loss, we first analyze trends within the structure of representations produced by standard supervised contrastive learning. To do this, we train a ResNet-50 \cite{he2016deep} model on Cifar-100 and then pass in the images associated with the test set $x_{test} \in X_{test}$. Every image is associated with the true class label $y_{gt}$ and the label that represents its taxonomy membership $y_{tax}$. Cifar-100  is constructed in such a way that there are 100 classes for $y_{gt}$ and 20 superclasses for $y_{tax}$. From the output of the network, we get the associated representation for each image $r_{i} \in R$. From this setup, we can analyze properties of the representation spaces produced by different subsets of the test set. 

\begin{figure}[ht]
\centering
\includegraphics[scale = .4]{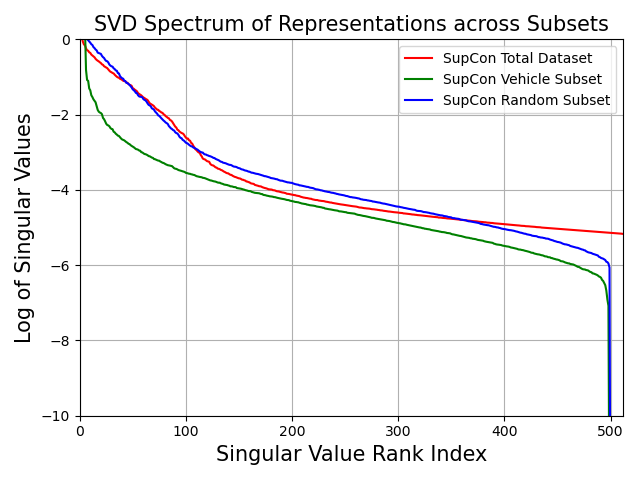}
\caption{This shows the SVD spectrum of the covariance matrix of subsets of the Cifar-100 test set from a model trained with SupCon \cite{khosla2020supervised}. We show the spectrum for the test set as a whole as well as for a subset consisting of images belonging to the vehicle superclass as well as an equally size random assortment of images.}
\label{fig: svd_spectrum}
\end{figure}

\begin{figure}[ht]
\centering
\includegraphics[scale = .35]{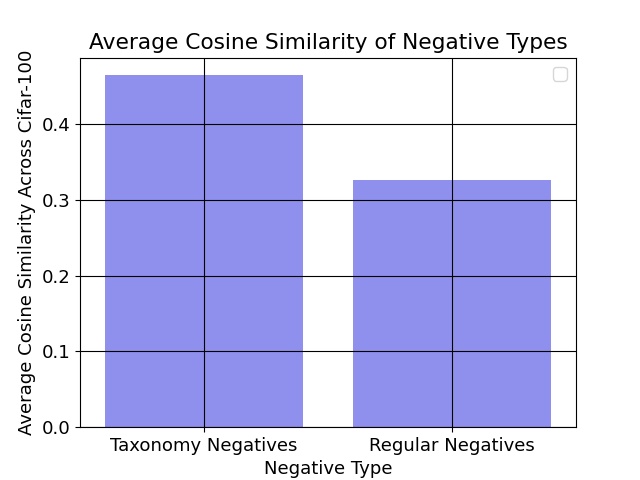}
\caption{This plot shows the average pairwise cosine similarity between the anchor image and taxonomy and regular negatives. }
\label{fig: superclass_distance}
\end{figure}

\begin{figure}[ht]
\centering
\includegraphics[scale = .25]{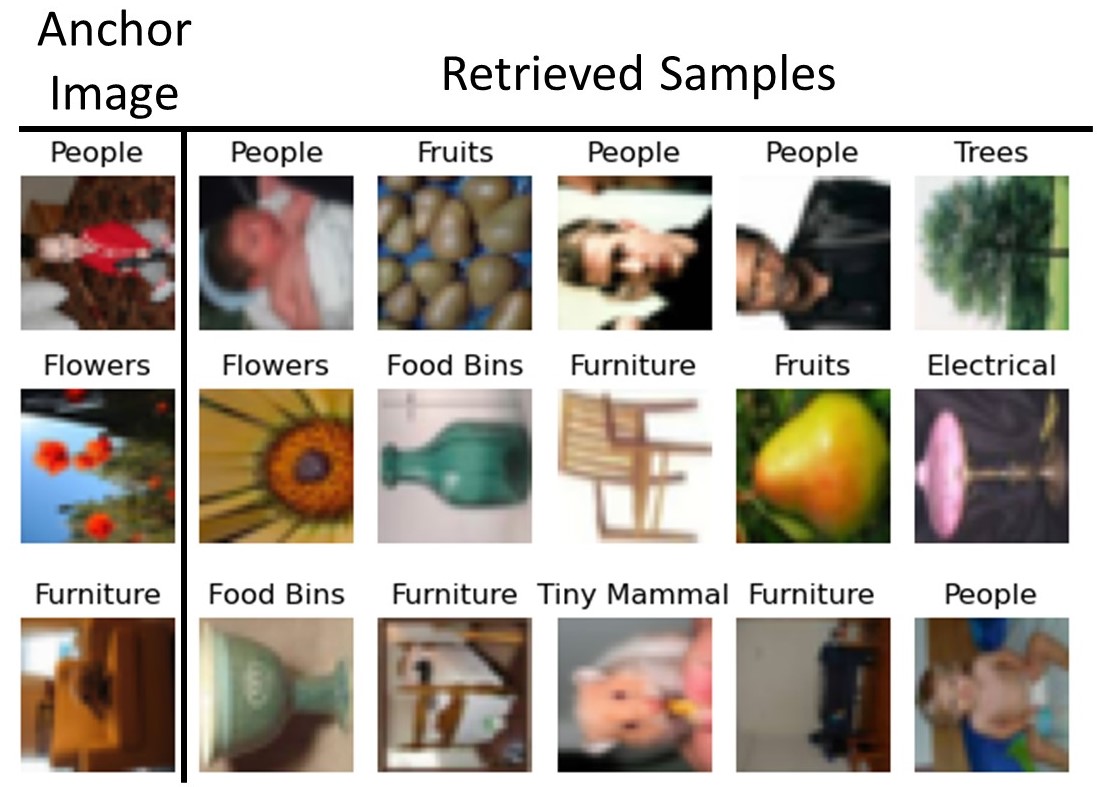}
\caption{For each anchor image from Cifar-100, we retrieve the images with the highest cosine similarity in the same batch. We show the taxonomic grouping label for each image. \vspace{-.5cm}}
\label{fig: retrieved}
\end{figure}

\begin{figure}[ht]
\centering
\includegraphics[scale = .4]{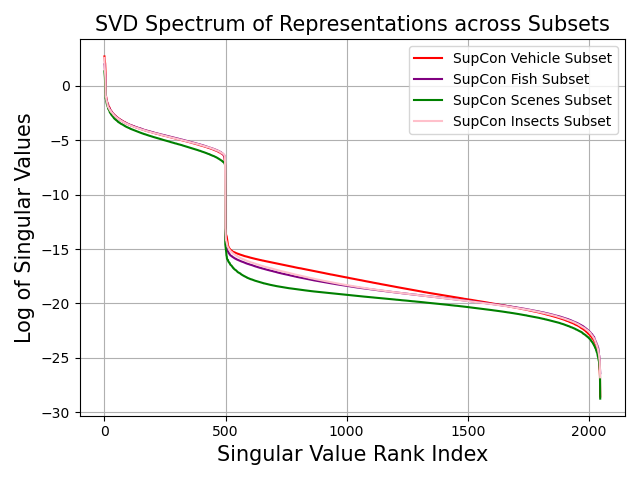}
\caption{This shows the SVD spectrum of the covariance matrix of subsets of the Cifar-100 test set from a model trained with SupCon \cite{khosla2020supervised}. We show the svd spectrum for specific superclass subsets. }
\label{fig: svd_spectrum_subclass}
\end{figure}

\begin{figure*}[ht]
\centering
\includegraphics[scale = .3]{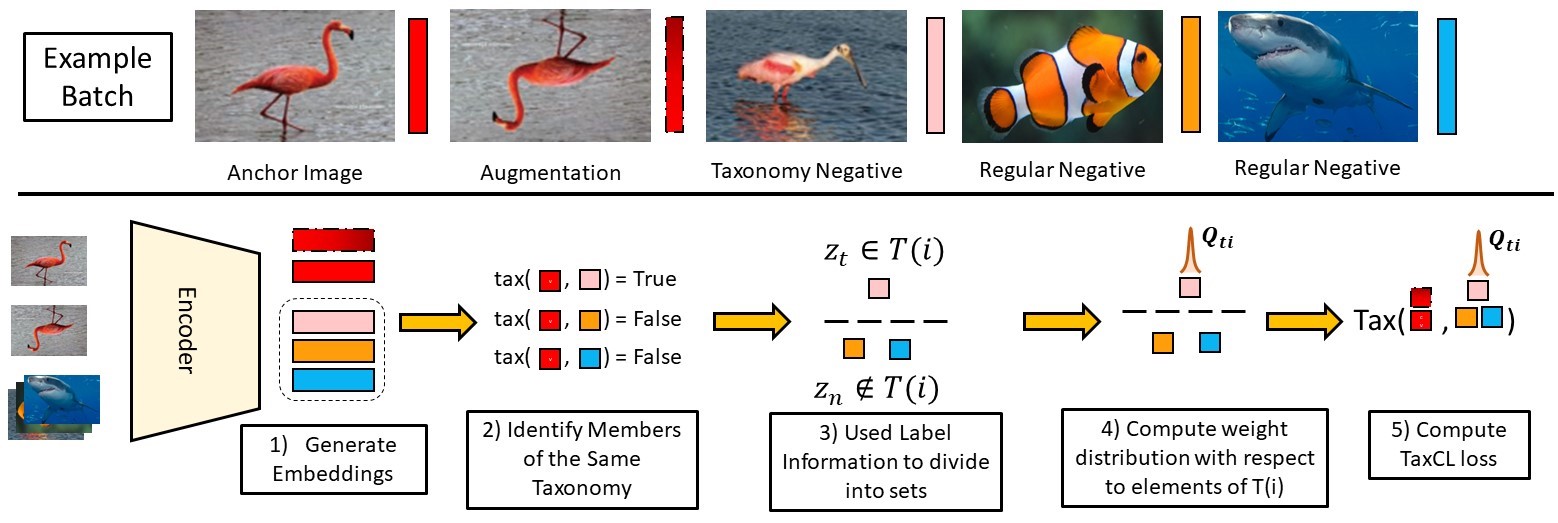}
\caption{This is the process by which the TaxCL loss is computed.}
\label{fig: main_visual}
\end{figure*}
One type of analysis relates to the dimensional collapse of the representations space. A core principal of contrastive learning is that elements of the positive set project closer to each other than that of the negative set. However, without proper design considerations, degenerate solutions such as mapping all positive points to the same location in space are possible. This situation represents complete dimensional collapse. Recent work \cite{garrido2023rankme,jing2021understanding} has shown that the degree of dimensional collapse correlates with the quality of the representation space and its corresponding performance on downstream tasks. We adapt this analytical point towards specific subsets of the dataset. In Figure \ref{fig: svd_spectrum} we show the SVD spectrum $U\lambda V^{T}$ of the covariance matrix $C = rr^{T}$ of the representations associated with different subsets of the Cifar-100 test set. We first show the spectrum of singular values of the test set as a whole which corresponds to the values of $\lambda$. This is less susceptible to dimensional collapse due to the wide distribution of data at its disposal. However, we also extract representations $r_{tax}$ for only images associated with a specific $y_{tax}$ (vehicle superclass) as well as an equally sized random subset of data. This results in a taxonomy specific covariance matrix $C_{tax} = r_{tax}r^{T}_{tax}$ as well as its own set of singular values $\lambda_{tax}$. We observe that while both covariance matrices exhibit a svd spectrum with rapidly declining singular values, the one drawn from a random subset maintains a higher magnitude in its highest ranking singular values as well as consistently larger singular values overall. This suggests  that samples drawn from the same taxonomy are more prone to dimensional collapse and this would correspondingly cause low separability between representations of elements from the same taxonomy as shown by Cover's Theorem \cite{cover1965geometrical}. As a consequence of this, we see in Figure \ref{fig: superclass_distance} on the Cifar-100 test set that the average pairwise cosine similarity between each anchor image representation $r_{i}$ and negatives that are members of the same taxonomy $r_{tax}$ is on average higher than negatives drawn from samples outside the same taxonomic grouping which we denote as regular negatives with an associated representation $r_{reg}$. This type of analysis has inspired approaches such as \cite{robinson2020contrastive} that attempt to use cosine similarity values to create a weighting on the negative distribution without explicitly accounting for taxonomic information. However, we show in Figure \ref{fig: retrieved} example images and their associated taxonomic grouping of the samples with the highest cosine similarity with the anchor image from a model trained with SupCon. We observe that the retrieved samples are oftentimes members of taxonomic groupings that differ from that of the anchor image. This means that methods that rely on weighting based on cosine similarity oftentimes fails to identify the semantic samples that have the most features in common with the anchor image and lose the advantages shown in part a) of Figure \ref{fig: batch_view}. Our method addresses this issue, by providing a loss that can directly integrate identified taxonomic relationships. 

Another interesting point is that the svd spectrum of different taxonomic groupings exhibits some variance. In other words, each grouping exhibits its own degree of dimensional collapse that is unique to itself. This also motivates treating taxonomy negatives separately because each independent taxonomic grouping has its own individual structure that must be accounted for.

\section{Methodology}

The standard supervised contrastive loss takes an input image $i$ to produce an output representation $r_i$ after passing through a network $f(.)$. This representation is used as input to a projection network $g(.)$ to produce the embedding $z_i$. The embeddings of its positive set $z_p$ are produced from the set of positives $p \in P(i)$. In this case, the set of positives are drawn from images with the same class label $y$ as image $i$. The set of positives and negatives is denoted as $a \in A(i)$ and its associated embedding is $z_a$. These embeddings are scaled by the temperature parameter $\tau$. Overall, this results in the loss: 

 $$
    L_{supcon} = \sum_{i\in{I}} \frac{-1}{|P(i)|}\sum_{p\in{P(i)}}log\frac{exp(z_{i}\cdot z_{p}/\tau)}{\sum_{a\in{A(i)}}exp(z_{i}\cdot z_{a}/\tau)}.
$$

In our TaxCL framework, our goal is to decompose the set of all negatives from $\sum_{a\in{A(i)}}exp(z_{i}\cdot z_{a}/\tau)$ to:

$$
Q_{ti}(\sum_{t\in{T(i)}}exp(z_{i}\cdot z_{t}/\tau)) + \sum_{n\notin{T(i)}}exp(z_{i}\cdot z_{n}/\tau)  
$$
where $T(i)$ is a subset of $A(i)$ that represents the taxonomic negatives in the batch with embeddings $z_t$ and $n$ represents the elements  that are not in $T(i)$ with an associated embedding $z_n$. $Q_{ti}$ represents a weighting function specific to the current batch. There are a variety of functions that could be used to appropriately scale the weighting of the taxonomic negatives. In this case, we use a function similar to the work of \cite{robinson2020contrastive}, but the difference is that its averaging only happens with respect to the elements belonging to $T(i)$ and no additional $\beta$ parameter is used. The overall construction of this function $Q_{ti}$ is to normalize the exponential of the cosine similarities of the taxonomic negatives by summing across this computed quantity and dividing by the mean of all taxonomic negatives. This results in the reweighted taxonomic negatives that are then added to the exponential of the cosine similarities of the positive set multiplied by the batch size and scaled by a $\tau_{+}$ parameter that is set to .1 for all experiments. The overall process for computing our loss is detailed visually in Figure \ref{fig: main_visual}. The associated loss is what we refer to as the TaxCL loss $L_{tax}$ and it has the form: 

\begin{equation}
\tiny
 \sum_{i\in{I}} \frac{-1}{|P(i)|}\sum_{p\in{P(i)}}log\frac{exp(z_{i}\cdot z_{p}/\tau)}{Q_{ti}(\sum_{t\in{T(i)}}exp(z_{i}\cdot z_{t}/\tau)) + \sum_{n\notin{T(i)}}exp(z_{i}\cdot z_{n}/\tau) } 
\label{eq:DD}
\end{equation}

We also formulate a loss that is a linear combination of both the supervised contrastive loss and our TaxCL loss with a weighting term $\alpha$. In this case, the taxonomy information adds additional structure by acting as regularization on top of the loss produced by the standard supervised contrastive loss. This takes the form:
$$
L_{combined} = (1-\alpha) L_{supcon} + \alpha L_{TaxCL}
$$

\section{Results Analysis}
\subsection{Training Setup}

We make use of three datasets in our analysis: Cifar-100, OLIVES \cite{prabhushankar2022olives}, and Cure-OR \cite{temel2017cure}. These datasets were chosen due to exhibiting different types of taxonomies as a natural function of the construction of each dataset. Lets assume each image $i$ in each dataset is associated with a label $y_{gt}$ and a taxonomy based label $y_{tax}$. Cifar-100 is a standard natural image dataset with $y_{gt}$ representing the fine-grained semantic class while $y_{tax}$ represents a superclass that represents some category of interest that 5 of the 100 classes in the dataset belong to. Cure-OR is a noise analysis dataset where the task is object recognition under the presence of noise. We set $y_{gt}$ as the class of each object and $y_{tax}$ as the associated noise type. OLIVES is a biomarker detection dataset within the setting of Optical Coherence Tomography (OCT) scans. $y_{gt}$ is the presence or absence of the diabetic macular edema biomarker, while $y_{tax}$ is the patient identity associated with a particular image. 

\begin{table}[]
\centering
\begin{tabular}{@{}cc@{}}
\toprule
\multicolumn{2}{c}{Performance on Cifar-100}        \\ \midrule
Method       & Performance      \\ \midrule
SupCon \cite{chen2020simple}      & 74.06\%       \\
SupHCL \cite{robinson2020contrastive}      &  73.56\%         \\
\midrule
TaxCL & \textbf{74.43\%} \\ 

SupCon + TaxCL Combined & \textbf{74.90\%} \\ \midrule

\end{tabular}
\caption{This shows the performance of TaxCL compared to other strategies on Cifar-100. }
\label{tab:loss_compare}
\end{table}
\begin{table}
\small
\centering
\begin{tabular}{@{}cccc@{}}
\toprule
\multicolumn{3}{c}{Analysis of SuperClass Decomposition}        \\ \midrule
Method           & OLIVES  & Cure-OR \\ \midrule
SupCon  \cite{khosla2020supervised}                         & 77.78\%   & 19.36\% \\
TaxCL             & 78.00\%     & 20.27\%    \\
SupCon + TaxCL               &  \textbf{80.12\%} & \textbf{27.61}\%\\ \bottomrule
\end{tabular}
\caption{This shows the performance of TaxCL in datasets with other types of taxonomies. \vspace{-.5cm}}
\label{tab:supervisedsuperclass}
\end{table}

All datasets were trained slightly differently depending on the features associated with each dataset. Each training setup is split into contrastive pre-training of the backbone ResNet-50 \cite{he2016deep} network followed by linear fine-tuning on top of frozen features from the backbone. For all datasets, we use random resized crop, color jitter, random grayscale, gaussian blur, and random horizontal flips as augmentations during contrastive pre-training. For Cifar-100, we train for 400 epochs with the LARS optimizer, a batch size of 256, temperature parameter of .2, learning rate of .4, and warmup cosine decay scheduler. For Cifar-100, we also compare our methods with the traditional supervised contrastive loss as well as a supervised version of the hard contrastive loss. For the supervised hard contrastive loss, we make use of the training parameters of their original paper. For Cure-OR, we trained the backbone for 200 epochs with a stochastic gradient optimizer, batch size of 256, a learning rate of .05, decay rate of .1, and decay epoch of 150. For OLIVES, we train for 25 epochs, with a batch size of 128, a stochastic gradient optimizer, and a learning rate of .05. For linear fine-tuning, we attach a linear layer and train for 100 epochs using stochastic gradient descent with a learning rate of .1 and a decay by a factor of 10 at epochs 60 and 80.

\subsection{Performance Analysis}

\begin{figure}[ht]
\centering
\includegraphics[scale = .45]{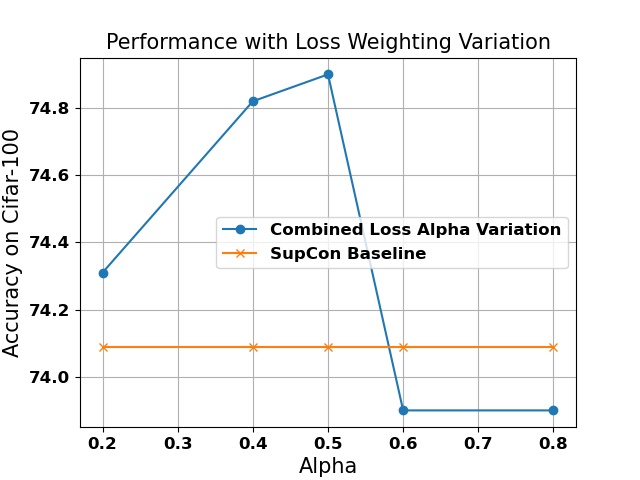}
\caption{This shows how performance varies in the combined loss as the alpha weighting parameter is varied. \vspace{-.5cm}}
\label{fig: loss_anal}
\end{figure}

\begin{figure}[ht]
\centering
\includegraphics[scale = .4]{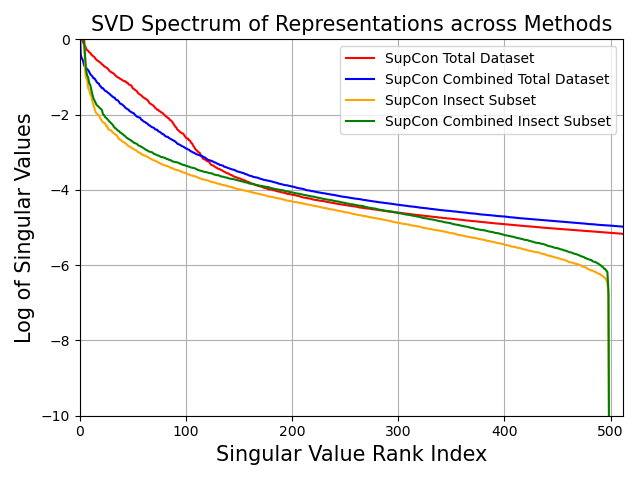}
\caption{This shows how the svd spectrum shifts between a standard trained SupCon model on Cifar-100 and a model trained using our TaxCL in a combined loss. We show the spectrum for the dataset as a whole as well as with respect to an exemplar superclass subset. \vspace{-.6cm}}
\label{fig: svd_spectrum_strategies}
\end{figure}

We first evaluate the performance of our method on Cifar-100 in Table \ref{tab:loss_compare}. We observe that both our TaxCL and Combined loss out-perform the standard SupCon training strategy as well as the supervised hard contrastive loss. Additionally, the combined loss out-performs all methods by a significant margin. These results indicate a variety of interesting trends that are a result of the manner in which each method was trained. For example, out-performing standard SupCon indicates that integrating an additional taxonomical structure improves the generalizability of the resulting representation space. However, the integration of this structure was best introduced as an additional regularization term on top of the standard supervised contrastive loss as shown by the significant performance gain when using TaxCL as part of a linear combination of losses with SupCon. Outperforming the supervised hard contrastive loss indicates that our method of decomposing the negatives into a taxonomy and regular set before computing a weighting term better reflects the semantic distribution of relationships in each batch. We also show in Figure \ref{fig: loss_anal} how performance varies as the alpha parameter in the combined loss varies. We observe that the best performance corresponds to an equal weighting between TaxCL and SupCon while an imbalanced weighting parameter results in worse performance for higher and lower $\alpha$ values. This improved performance can also be understood intuitively from an analysis of the representation space between traditional SupCon and our TaxCL and SupCon combined loss. In Figure \ref{fig: svd_spectrum_strategies}, we analyze the SVD spectrum of the covariance matrix of the test set representations for each method. We observe that out combined loss maintains a higher magnitude of singular values for more of its highest ranking singular values compared to SupCon on either the test set as a whole or our exemplar superclass subset. This indicates that our method partially reduces the dimensional collapse of samples belonging to the same taxonomic category.

We also show that our method generalizes well to other domains. In Table 2, we observe that in the medical setting of OLIVES and the noise object recognition setting of Cure-OR our method out-performs conventional SupCon. This performance improvement significantly increases when using SupCon and TaxCL in a combined loss. This shows that there exists  a variety of different types of information that can be useful for enforcing a taxonomical structure in the representation space. Additionally, the larger performance improvement in these settings compared to Cifar-100 indicates that this taxonomical semantic alignment matters more when real-world data complications exist such as class imbalance in the medical case or corruptions as in the case of the Cure-OR dataset.

\section{Conclusion}

In this work, we introduce a new supervised contrastive loss that allows direct integration of semantic taxonomy information during the representation learning process. Namely, we show how decomposing the distribution of negatives via higher order semantic labels can effectively result in a set of negatives that better approximates the relationships that exist within a batch. Additionally, we show that this strategy out-performs conventional supervised contrastive learning across a variety of datasets and improves performance further when used as a linear combination with the supervised contrastive loss. We also show through our experiments on other datasets that the notion of a taxonomy can be generalized in such a way that it can be adapted to the domain of interest.

\bibliographystyle{IEEEbib}
\bibliography{refs}

\end{document}